# A Survey of League Championship Algorithm: Prospects and Challenges

**Shafi'i Muhammad Abdulhamid[1*], Muhammad Shafie Abd Latiff[1], Syed Hamid Hussain Madni[1] and Osho Oluwafemi[2]**

[1]Faculty of Computing, Universiti Teknologi Malaysia, 81310 Skudai, Johor, Malaysia;
shafii.abdulhamid@futminna.edu.ng, shafie@utm.my, madni4all@yahoo.com
[2]Department of Cyber Security Science, Federal University of Technology, Minna, Nigeria;
femi.osho@futminna.edu.ng

## Abstract

The League Championship Algorithm (LCA) is sport-inspired optimization algorithm that was introduced by Ali Husseinzadeh Kashan in the year 2009. It has since drawn enormous interest among the researchers because of its potential efficiency in solving many optimization problems and real-world applications. The LCA has also shown great potentials in solving non-deterministic polynomial time (NP-complete) problems. This survey presents a brief synopsis of the LCA literatures in peer-reviewed journals, conferences and book chapters. These research articles are then categorized according to indexing in the major academic databases (Web of Science, Scopus, IEEE Xplore and the Google Scholar). The analysis was also done to explore the prospects and the challenges of the algorithm and its acceptability among researchers. This systematic categorization can be used as a basis for future studies.

**Keywords:** Global Optimization, League Championship Algorithm, NP-Complete Problems, Sport- Inspired Algorithm, Swarm Algorithms, Optimization Algorithm

## 1. Introduction

League Championship Algorithm (LCA) is a population based algorithmic framework for global optimization over a continuous search space first proposed by Kashan[1]. It is a Swarm optimization algorithm[2–5]. A general characteristic between all population based optimization algorithms similar to the LCA is that they both try to improve a population of achievable solutions to potential areas of the search space when seeking the optimization. LCA is a newly proposed stochastic population based algorithm for continuous global optimization which tries to imitate a championship situation where synthetic football clubs participate in an artificial league for a number of weeks. This algorithm has been tested in many areas and performed creditably well as compared to other known optimization schemes or heuristics algorithms[6].

Optimization algorithms are dynamic research areas with rapidly increasing literatures. It is a common knowledge among researchers that combinatorial optimization problems like the travelling salesman problem are non-deterministic polynomial times (NP-hard) problems[7,8], which implies that there are no fast solutions or exact solutions for such problems. Some of the most effective solutions to NP-hard problems are to utilize heuristic or metaheuristic algorithms. Heuristic algorithms are to look for solutions using trial and error, while the metaheuristic algorithm can be seen as a higher-level algorithm that use information and selection of the solutions to guide optimization problems with the intention of finding the optimality, even though such optimality is not mostly attainable. But many discrete and continues optimization problems are NP-complete, optimal solution techniques which we can execute as a polynomial time is also not attainable with a very high probability. Therefore, these types of problems are tackled by the construction of heuristics, metaheuristics or approximation algorithms which run in polynomial time and are often the best known method to find efficient solutions[9,10].

*Author for correspondence

In recent times, many inspirational optimization algorithms have gained large popularity among researchers in solving artificial intelligence, machine learning, computational intelligence, engineering applications, distributed computing (parallel, grid and cloud) and data mining. These includes; nature-inspired optimization algorithms[11], bio-inspired optimization algorithms[12], evolutionary algorithms[13], physics-inspired optimization algorithms[14], ecologically-inspired optimization algorithms[15] and now sports-inspired optimization algorithms[6]. The LCA is a sport-inspired optimization algorithm which is also a type of swarm algorithm. It has been used in several researches in comparison with other established techniques to evaluate its efficiency and in most cases the results obtained are quite remarkable.

The aim of this paper is to survey all available research articles on the League Championship Algorithm (LCA) and its applications. We also wish to highlight some potential application areas that are yet to be exploited and some challenges. Therefore, the paper is organised as follows: Section 2 presents the LCA as it appears in previous literatures. Section 3 reviews in brief, a synopsis of the LCA researches. Section 4 investigates the acceptability of the LCA algorithm within the academic community. Section 5 presents the future prospects of using the LCA and its applications in other research areas. Section 6 describes some of the challenges faced by the users of the LCA algorithms and section 7 is the conclusion of the survey.

## 2. League Championship Algorithm in Literatures

Since the introduction of LCA in the year 2009, different researchers tried to apply it in different areas of research to solve specific problems. This section discussed the chronological history of this sport inspired algorithm as the researches continue to appear in major academic indexing and abstracting databases.

### 2.1 LCA in Major Academic Databases

Four major academic databases were used in this survey (in no particular order) to check the presence of the papers published with the LCA optimization technique. This gives us more inside on how the research community is utilizing this algorithm by its present in the major academic databases for indexing and abstracting. The research databases used for this survey are; the ISI (Thomson Reuters) also called the Web of Science, the Scopus, the IEEE Xplore and the Google Scholar.

Table 1 shows a record of research articles that used the LCA in the selected databases as at 31st January, 2015. It also shows that the LCA have a very high presence in the Google Scholar academic database as at the period under review as almost all the LCA research papers are indexed in it. The Scopus academic database powered by the Elsevier has the second largest indexed documents of the LCA research articles. It contains some articles as at the period under review (17, 20, 24, 6, 18, 19, 1) etc. The IEEE Xplore database contains three research documents that implemented the LCA optimization technique (17, 19, 1). The Web of Science which is the most important scientific database contains about five documents (6, 18) etc. as at the time under review[24]. The fact that the LCA only made it to the Web of Science after just about four years shows that the algorithm is not get the desired attention that it deserved.

## 3. A Synopsis of LCA

The research papers presented in this synopsis are gathered through searches from the above mentioned academic indexing databases. These same papers are also available through http://drkashan.ir. It is also important to mention that, this survey only considers publications written in English, as we are aware that there are some LCA publications written in other languages especially Arabic.

From Figure 1, Kashan[16] presented a paper that proposed and introduced a new evolutionary algorithm called League Championship Algorithm (LCA) for global optimization, which mimics the sport league championships. It is a new algorithm for numerical function optimization. Kashan and Karimi[17] tasted the effectiveness of the proposed optimization algorithm by measuring the test functions from a recognized yardstick, usually adopted to authenticate new constraint-handling algorithms strategy. The outcome derived from the proposed technique are very competitive with regards to other well-known techniques that already exist for constrained optimization and testify that the new algorithm can be regarded as an capable technique for optimization in the presence of constraints.

Kashan[18] modified and adapted LCA for constrained optimization in mechanical engineering design. The algorithm was used for a practicability criterion to favour the search toward feasible regions is included in addition the objective value condition. It then generates many children which increase the possibility of an entity to generate a better result. A multiplicity system is also adopted, which permits infeasible solutions with a potential objective value to come head of the feasible solutions. The effectiveness of LCA system was measured up against other similar algorithms on benchmark problems where the experimental outcome indicates that LCA is a very viable algorithm. The effectiveness of the LCA system was also measured on well-known mechanical design problems and outcomes are evaluated with the outcomes of 21 other constrained optimization algorithm techniques. The outcome of this comparison shows that with a smaller number of evaluations, LCA guarantees discovering the true optimum of these problems.

**Table 1.** LCA in Major Academic Indexed Databases

| Reference | Application Area | ISI | Scopus | IEEE Xplore | Google/GoogleScholar |
|---|---|---|---|---|---|
| Kashan (16) | LCA Introduced | √ | √ | √ | √ |
| Kashan and Karimi (17) | LCA Refined | | √ | √ | √ |
| Kashan (18) | Mechanical Engineering Design | √ | √ | | √ |
| Kashan, Karimiyan (19) | Modified LCA for "between two halves analysis" | | √ | √ | √ |
| Pourali and Aminnayeri (20) | Minimizing Earliness/Tardiness Penalties | √ | √ | | √ |
| Kejani (21) | Reliability Optimization | | | | √ |
| Lenin, Reddy (22) | Reactive Power Dispatch Problem | | | | √ |
| Stephen and PVGD (23) | Simple LCA | | | | √ |
| Sun, Wang (24) | Resource Allocation Mechanism for Cloud | | √ | | √ |
| Edraki (25) | Engineering Design Optimization in Centrifuge Pumps | | | | √ |
| Kashan (6) | LCA for global Optimization | √ | √ | | √ |
| Kahledan (7) | Travelling Salesman Problem | | | | √ |
| Sebastián and Isabel (26) | Scheduling Job Shop Configuration | | | | √ |
| Bouchekara, Abdallh (27) | Optimization of Electromagnetic Devices | | | | √ |
| Bouchekara, Abido (28) | Optimal power flow | √ | √ | | √ |
| Seyed Mojtaba Sajadi, Ali Husseinzadeh Kashan (29) | Permutation Flow-Shop Scheduling Problem | | | | √ |
| Abdulhamid, Abd Latiff (30) Abdulhamid and Latiff (31) | Job Scheduling in IaaS Cloud Computing | | | √ | √ |

Kashan, Karimiyan[19] modifies the LCA system to solve numerical function optimization through the artificial modelling of the "Between Two Halves Analysis". The research worker tried to enhance the basic algorithm by modelling a between two halves like analysis beside the post-match SWOT (strengths/weaknesses/opportunities/threats) analysis to produce new outcomes. The result obtains from the modified algorithm was tested and compared with that of basic version and also the Particle Swarm Optimization Algorithm (PSO) on finding the global minimum of a number of benchmark functions. The outcome shows that the improved algorithm called RLCA, perform well in terms of both final solution quality and convergence speed.

Pourali and Aminnayeri[20] develops a new evolutionary LCA to tackle a new single machine scheduling nonlinear problem in Just-In-Time (JIT) system with batch delivery cost and different due dates. Regardless of its complexity, finding a solution to a non-convex function which reduces earliness and tardiness costs concurrently appears to be very helpful and practical in industry. In order to find a solution to this composite problem, the research presented a mathematical model and then designed a new discrete nonlinear LCA algorithm that was very proficient and helpful in combinatorial problems, in terms of computational time or solution quality.

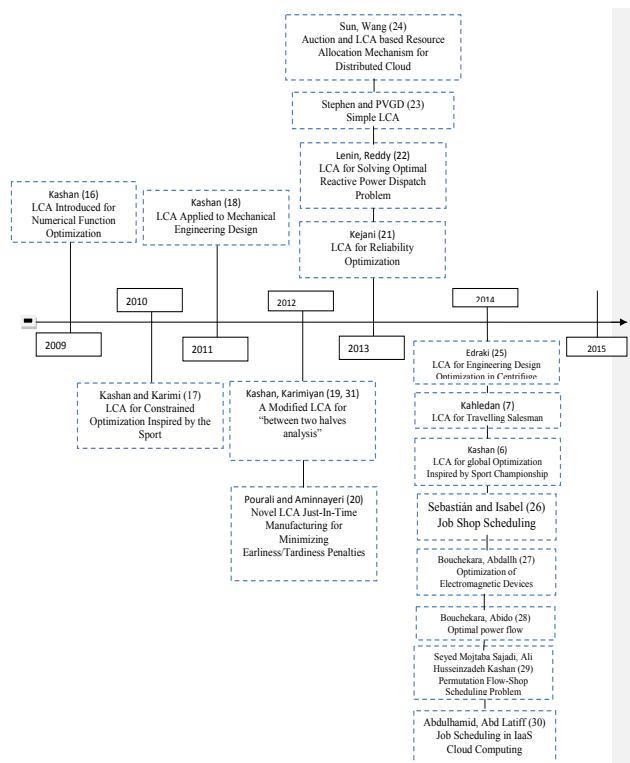

**Figure 1.** LCA in Literatures.

Kejani[21] came up with a new approach for reliability optimization based on LCA, while Lenin, Reddy[22] came up with another adapted LCA for solving optimal multi-objective reactive power dispatch problem. Then a modal analysis was used for fixed voltage stability estimation. The optimization of voltage stability margin is taken as the goals. The research shows that the LCA produce very good results. Stephen and PVGD[23] presented a simple LCA in an effort to make optimization process free from ambiguities. The new method provided a number of theoretical frameworks to utilize multiple optimization techniques simultaneously in a single optimization problem. This optimization process was been implemented to provide solution to image enhancement problem in a fingerprint, with some experimental results to authenticate the validity of this new process.

Sun, Wang[24] presents an auction and LCA-based resource allocation mechanism for distributed cloud computing. The research presented a double combinatorial auction derived from allocation system which was also derived from the properties of cloud computing resources and inspired by the elasticity and efficiency of microeconomic methods. The feedback assessment derived from reputation system with reduced coefficient of time and the chain of command of clients that was introduced was implemented to steer clear of malicious actions. To ensure informed scientific conclusions, the researchers came up with a price decision system derived from a back propagation neural network. In the price decision system, the different parameters are considered, so the requesting price can adapt to the varying supply-demand relation in the system. In view of the fact that the winner determination is an NP-complete problem, LCA was utilized to attain optimal allocation with the optimization objectives as market surplus and total reputation. The research was concluded by conducting empirical studies to show the practicability and efficiency of the anticipated mechanism.

Edraki[25] presents a new approach for engineering design optimization of centrifuge pumps based on LCA. Also Kahledan[7] applied the League Championship Algorithm to solve the Travelling Salesman Problem. The travelling salesman problem (TSP) tries to optimize a list of cities and the distances between each pair of cities, and then find the shortest feasible route that visits each city exactly once and returns to the origin city. It is also an NP-hard problem in combinatorial optimization, which is vital in operations research and theoretical computer science. The result of the LCA shows that its very effective compared to other known methods.

Kashan[6] comes up with an improved version of the LCA detailing the workings and functionalities of the different add-on modules of the algorithm. This latest paper also details more about the iterations, fitness value, generation of new solutions and stopping conditions. Lastly, an analysis was carried out to validate the foundation of the algorithm and the appropriateness of the updating equations collectively with investigating the consequence of diverse settings for the control parameters are carried out empirically on several of benchmark functions. The outcome of the analysis shows that the LCA shows potential performance suggesting that its further developments and practical applications would be worth investigating in the future research and applications.

Sebastián and Isabel[26] presents an implementation of the LCA in a Job Shop scheduling in an industrial situation. The Job Shop is a production system problem of "n" machines and "m" jobs to be distributed among the machines; each job is to follow particular production line and also need to use all machines. The optimum scheduling method need to be found in order to minimizing the production time and the make span. The modified algorithm was evaluated using instances of tests taken from the previously proposed methods in the literatures. The results shows the LCA produced more accurate values as compared with the previous methods.

From the synopsis of LCA literatures presented above, it shows that the algorithm has the flexibility, capability and efficiency of being adapted or adopted to solve different types of problems in different non-deterministic polynomial time (NP-hard or NP-complete) situations. As all the previous researches also applied the scheme to solve the NP-complete problems[32]. The most distinguished attribute of NP-complete problems is that no quick answer to them is known and also no exact solution is known. NP-complete problems are often addressed by using heuristic methods, evolutionary algorithms, and approximation algorithms[33–5].

## 4. Acceptability of LCA

In this section, we investigated how acceptable the LCA algorithm was over the years within the research community. By making simple search in some of the major academic databases revealed that the number of research documents (both research articles and documents citing LCA articles) that cited the terms "League Championship Algorithm" and "Champions League Algorithm" are increasing yearly. Below we presented the outcome of such investigation in the Google Scholar, Scopus, IEEE Xplore and also in the Web of Science. The Google Scholar is a very robust open access academic database. It keeps record of many scientific publications ranging from research articles, books, journals and conferences. These records are normally obtained from different sources. It is also a very reliable search engine that keeps records of citations, $h$-index and i10-index.

The Scopus academic database is one of the most reliable scientific research databases. But, unlike the Google Scholar it is not an open access, as researchers needs to subscribe to get access to the facilities. It is a very versatile indexing facility with the capabilities of graphical analysis of articles and cited documents. The IEEE Xplore is also an academic database that hosts millions of academic researches worldwide. It is known for archiving most science and technology journals, conferences and transactions.

Web of Science is also called the ISI or the web of knowledge. It is an online subscription-based scientific citation indexing service controlled by Thomson Reuters that presents a detailed citation search and impact factor of academic journals. It gives access to multiple databases that reference cross-disciplinary research, which allows for in-depth exploration of specialized sub-fields within an academic or scientific discipline.

If you consider that the year 2015 is just one month old as at the time of this research. Figure 2 shows that the number of documents that contains the terms "League Championship Algorithm" or "Champions League Algorithm" in the Google Scholar database is increasing yearly. This implies that, the prospect of the LCA algorithm amongst the research community is high and the algorithm is also gaining more acceptances. This means that the LCA is an effective scheme for solving NP-complete problems. The LCA also got a very important citation through the Google Scholar from the yearly soft computing Clever Algorithm book called Innovative Computational Intelligence[36]. This shows that, the LCA was well received within the research community.

Figure 2 also shows that similarly the number of documents that contains the terms "League Championship Algorithm" or "Champions League Algorithm" in the Scopus database is also increasing yearly. This implies that, the prospect of the LCA algorithm amongst the research community is high and the algorithm is also gaining more acceptances. This means that, the LCA is an effective scheme for solving NP-complete problems. Similar to the investigation in the Google Scholar, the result shows that the algorithm was fairly utilized in many areas of scientific research.

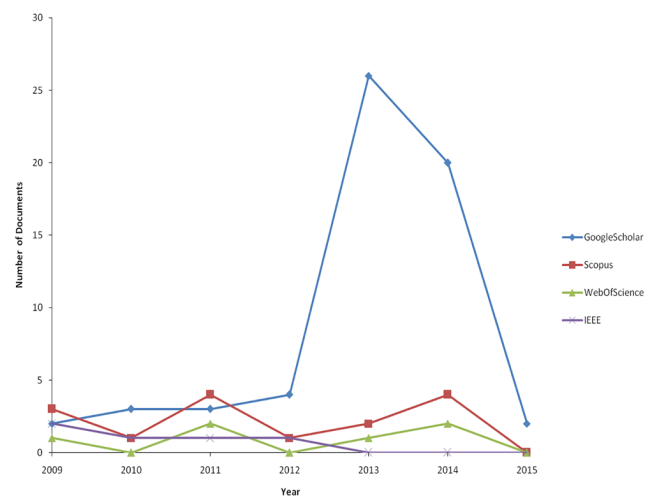

**Figure 2.** Documents Citing LCA.

The graph above also shows that both IEEE Xplore and the Web of Science recorded both citations and archived documents containing the terms "League Championship Algorithm" or "Champions League Algorithm". Even though the number of documents in these databases is not much, but it shows great promises of growing over time. This implies that, the algorithm under review is well accepted by the academic communities.

## 5. The Prospects of LCA Optimization

The league championship algorithm is still an evolving soft computing algorithm that is gradually gaining relevance every passing day within the research community. The prospect of using the LCA in other application areas is very high. The performance of the scheme in previous literatures implies that the algorithm can also be widely applied in other similar research areas, especially for solving NP-hard problems. Table 2 presented some applications that are yet to be explored with some associated research problems.

From Table 2, it shows that the LCA as an affective algorithm used in previous researches to tackle NP-complete problems can as well effective in a wide range of research domains. This includes but not limited to; data mining, big data, mobile computing, sensor networks, engineering design, distributed systems, graph colouring, machine learning, timetable, subset problem, etc. The research issue in the different domains also differs greatly, ranges from resource allocation, search techniques, energy conservation, routing issues, assignment problem to mention but a few.

## 6. The Challenges of LCA Optimization

Even though the league championship optimization algorithm has made great progresses in recent years, there are some essential and attractive challenges that needs to be probing further. Some of these problems are listed below.

### 6.1 Generalization

Generalization is a very important feature of any optimization algorithm and this also includes the LCA. Evaluating the LCA's generalization can be a complex task because of the diverse implementations. The complexity is correlated to that of evaluating the generalization ability of LCA system in general. It is not unusual to come across researches that only relay a algorithm developed at a certain number of generations. But it is not normally clear, if such an algorithm is the outcome of a single run or the mean of many runs. More so, in some cases it is not very clear how to choose when to discontinue getting a good system. It is also uncertain how fit the LCA might generalize to diverse situations in these cases[83].

**Table 2.** Future Application Areas

| Application Domain | Problems |
|---|---|
| Big Data and Data Mining | Search technique[37], resource allocation[38], resource management[39], etc. |
| Mobile Computing and Sensor Networks | Resource allocation[40], energy conservation problem[41], resource distribution[42], search technique[43], etc. |
| Engineering Design | Toy problem[44,45], chaotic sequences[46], Cognitive engineering design problem[47], structural design problem[48], etc. |
| Networks and Distributed (i.e. parallel, grid and cloud) Systems | Routing problem[49,50], assignment problem[40,51], resource allocation[52] and resource distribution[53], etc. |
| Graph Colouring | Assignment problem[54,55]. |
| Resource Constrained Projects | Resource allocation[56,57], resource distribution[58] and scheduling[59]. |
| Machine Learning | Learning of classification rules[60,61], Learning the Structure of Bayesian Networks[62], learning rules in a fuzzy system[63] and quadratic assignment problem[64]. |
| University Timetable | Resources allocation[65] and scheduling[66,67]. |
| Subset Problems | Multiple knapsack problem[68–70], maximum independent set problem[71–73], maximum clique problem[74,75] and redundancy allocation problem[76,77], etc. |
| Other Applications | Open shop[78] and group shop scheduling[79], permutation flow shop problem[80], set covering problem[81], 2D-HP Protein Folding[82], etc. |

Most of the related LCA literatures show that some enhancements need to be done to suit a particular situation. Different problems have different conditions and parameters, and therefore may need different approach to solve. This problem of generalization is not particular to the LCA as most optimization and heuristic algorithms also suffered from this issue. In essence, more comprehensive studies of generalization and standardization in LCA will greatly improve it applicability in other areas.

### 6.2 Robustness

Robustness is an important feature for the reliability of any optimization scheme. In case of failures in an optimization scheme, the failure detection algorithm should instantaneously identify the error and starts a reprogramming that correct the process. In the LCA, such a scheme is not clearly spelt out, and therefore failure and reliability that can leads to robustness is an open issue till now. However, the robustness of LCA scheme may be accomplished through many methods of failure identification and restore, and the robustness of LCA quality should be further scrutinized in future researches.

### 6.3 Hybridization

Hybridization is acknowledged these days to be an important part of optimization algorithms. Most of the established heuristics and optimization schemes especially in computation intelligence (CI) include components and ideas derived from other optimization schemes. Even though, the LCA also derived some of its features from other swam algorithms, hybridization with other optimization algorithms (both heuristics and CI) need to be explore the more to increase its versatility and utility. However, these hybridizations most at times reach their limits when either large-scale problem instances with massive search spaces or highly constrained problems for which it is complex to get feasible solutions are considered[84]. Therefore, researchers are encouraged to investigate the integration of more classical CI, heuristics and operation research (OR) techniques into the LCA scheme. But, one reason why the LCA algorithm is especially suited for hybridization is its constructive nature.

### 7. Conclusion

In this survey paper, we gave a briefed description of the origins of the LCA algorithm. Then, we outlined the synopsis of some of the existing theoretical and experimental results in LCA literature. We provided a survey on a very interesting recent research direction as the prospects of future LCAs. The acceptability of this sport-inspired technique was investigated using some of the major academic databases. The paper also pointed out some of the challenges of the algorithm which includes; generalization, robustness and the hybridization of LCA algorithms with more classical CI, heuristics and operations research schemes. In conclusion, haven studied the LCA and most researches that have in one way or the other used the algorithm to solve problems in similar application areas, and based on the chronological literatures detailed in this paper, it shows that the LCA is a global optimization algorithm that performed superbly and efficiently in solving many NP-hard problems and this research direction offers many possibilities for valuable future research.

### 8. Acknowledgment

The first author would like to express his appreciation for the support of Universiti Teknologi Malaysia (UTM) International Doctoral Fellowship (IDF), while the second author would like to express his gratitude for the UTM, Research University Grand Q. J130000.2528.05H87 sponsorship for this research. We also wish to appreciate Dr. Kashan A. H. for his assistance on research materials.

### 9. References

1. Kashan HA, League championship algorithm: a new algorithm for numerical function optimization. Proceedings of IEEE International Conference on Soft Computing and Pattern Recognition (2009 SOCPAR'09); 2009. p. 43–8.
2. Fourie P, Groenwold A. The particle swarm optimization algorithm in size and shape optimization. Structural and Multidisciplinary Optimization. 2002; 23(4):259–67.
3. Trelea IC. The particle swarm optimization algorithm: convergence analysis and parameter selection. Inform Process Lett. 2003; 85(6):317–25.

4. Jiang Y, Hu T, Huang C, Wu X. An improved particle swarm optimization algorithm. Appl Math Comput. 2007; 193(1):231–9.
5. Chen C-Y, Ye F. Particle swarm optimization algorithm and its application to clustering analysis. IEEE International Conference on Networking, Sensing and Control; 2004. p. 789–94.
6. Kashan HA. League Championship Algorithm (LCA): an algorithm for global optimization inspired by sport championships. Applied Soft Computing. 2014; 16:171–200.
7. Kahledan S. A League Championship Algorithm for Travelling Salesman Problem. Iran: Najaf Abad Branch, Azad University; 2014.
8. Pappalardo E, Ozkok BA, Pardalos PM. Combinatorial Optimization Algorithms. Handbook of Combinatorial Optimization. Springer; 2013. p. 559–93.
9. Liu X, Yan W, Xun S, Qiang T. Solution of Aircraft Trim Problem Based on Genetic Algorithms. Proceedings of the 9th International Symposium on Linear Drives for Industry Applications; 2014.
10. Tsai P-W, Istanda V. Review on cat swarm optimization algorithms. 3rd International Conference of IEEE on Consumer Electronics, Communications and Networks (CECNet); 2013.
11. Yang X-S. Engineering optimizations via nature-inspired virtual bee algorithms. Artificial intelligence and knowledge engineering applications: a bioinspired approach. Springer; 2005. p. 317–23.
12. Gandomi AH, Alavi AH. Krill herd: A new bio-inspired optimization algorithm. Comm Nonlinear Sci Numer Simulat. 2012; 17(12):4831–45.
13. Kavousi-Fard A, Samet H. Multi-objective performance management of the capacitor allocation problem in distributed system based on adaptive modified honey bee mating optimization evolutionary algorithm. Elec Power Compon Syst. 2013; 41(13):1223–47.
14. Biswas A, Mishra K, Tiwari S, Misra A. Physics-inspired optimization algorithms: a survey. Journal of Optimization. 2013.
15. An J, Kang Q, Wang L, Wu Q. Mussels wandering optimization: an ecologically inspired algorithm for global optimization. Cognitive Computation. 2013; 5(2):188–99.
16. Kashan AH. League championship algorithm: a new algorithm for numerical function optimization. Proceedings of IEEE International Conference on Soft Computing and Pattern Recognition (SOCPAR'09); 2009.
17. Kashan HA, Karimi B. A new algorithm for constrained optimization inspired by the sport league championships. 2010 IEEE Congress on Evolutionary Computation (CEC); 2010; IEEE.
18. Kashan HA. An efficient algorithm for constrained global optimization and application to mechanical engineering design: League championship algorithm (LCA). Computer-Aided Design. 2011; 43(12):1769–92.
19. Kashan HA, Karimiyan S, Karimiyan M, Kashan MH. A modified league championship algorithm for numerical function optimization via artificial modelling of the between two halves analysis. 2012 Joint 6th International Conference on Soft Computing and Intelligent Systems (SCIS). 13th International Symposium on Advanced Intelligent Systems (ISIS); 2012; IEEE.
20. Pourali Z, Aminnayeri M. A novel discrete league championship algorithm for minimizing earliness/tardiness penalties with distinct due dates and batch delivery consideration. Advanced Intelligent Computing: Springer; 2012. p. 139–46.
21. Kejani T. A new approach for reliability optimization based on league championship algorithm (LCA). Iran: Najaf Abad Branch, Azad University; 2013.
22. Lenin K, Reddy BR, Kalavathi DMS. League Championship Algorithm (LCA) for Solving Optimal Reactive Power Dispatch Problem. International Journal of Computer and Information Technologies (IJOCIT). 2013; 1(1):1–19.
23. Stephen MJ, PVGD PR. Simple League Championship Algorithm. Int J Comput Appl. 2013; 75(6):28–32.
24. Sun J, Wang X, Li K, Wu C, Huang M, Wang X. An auction and league championship algorithm based resource allocation mechanism for distributed cloud. Advanced Parallel Processing Technologies. Springer; 2013. p. 334–46.
25. Edraki S. A new approach for engineering design optimization of centrifuge pumps based on league championship algorithm. Iran: Science and Research Branch, Azad University; 2014.
26. Sebastian AR, Isabel LR. Scheduling to job shop configuration minimizing the make span using champions league algorithm. Fray Ismael Leonardo Ballesteros Guerrero OP, editor. Tunja: Division de Arquitectura e Ingenierias, Universidad Santo; 2014.
27. Bouchekara H, Abdallh A, Hamza Kherrab LD, Mehasni R. Design optimization of electromagnetic devices using the League Championship Algorithm. International Workshops on Optimization and Inverse Problems in Electromagnetism (OIPE); 2014.
28. Bouchekara H, Abido M, Chaib A, Mehasni R. Optimal power flow using the league championship algorithm: a case study of the Algerian power system. Energ Convers Manag. 2014; 87:58–70.
29. Seyed Mojtaba Sajadi, Ali Husseinzadeh Kashan, Khaledan S. A new approach for permutation flow-shop scheduling problem using League Championship Algorithm. Proceedings of CIE44 and IMSS'14; 2014.

30. Abdulhamid SM, Abd Latiff MS, Ismaila I. Tasks scheduling technique using league championship algorithm for make span minimization in IAAS cloud. ARPN Journal of Engineering and Applied Sciences. 2014; 9(12):2528–33.
31. Abdulhamid SM, Abd Latiff MS. League Championship Algorithm based job scheduling scheme for infrastructure as a service cloud. 5th International Graduate Conference on Engineering, Science and Humanities (IGCESH 2014). Johor Bahru, Malaysia: Universiti Teknologi Malaysia; 2014.
32. Ullman JD. NP-complete scheduling problems. J Comput Syst Sci. 1975; 10(3):384–93.
33. Leyton-Brown K, Hoos HH, Hutter F, Xu L. Understanding the empirical hardness of NP-complete problems. Communications of the ACM. 2014; 57(5):98–107.
34. Abdulhamid SM, Abd Latiff MS, Bashir MB. Scheduling techniques in on-demand grid as a service cloud: a review. Journal of Theoretical and Applied Information Technology. 2014; 63(1).
35. Wu K, de Abajo JG, Soci C, Shum PP, Zheludev NI. An optical fiber network oracle for NP-complete problems. Light: Science and Applications. 2014; 3(2):e147.
36. Xing B, Gao W-J. Introduction to Computational Intelligence. Innovative Computational Intelligence: a rough guide to 134 clever algorithms. Springer; 2014. p. 3–17.
37. Bashir MB, Abd Latiff SM, Atahar AA, Yousif A, Eltayeeb ME. Content-based Information Retrieval Techniques Based on Grid Computing: A Review. IETE Technical Review. 2013; 30(3).
38. Skillicorn D, Talia D. Mining large data sets on grids: issues and prospects. Comput Informat. 2012; 21(4):347–62.
39. Kantardzic M. Data mining: concepts, models, methods, and algorithms. John Wiley and Sons; 2011.
40. Hyytia E, Spyropoulos T, Ott J. Optimizing offloading strategies in mobile cloud computing. 2013.
41. Yuan Q, Zhang R, Chu F, Dai W, editors. ECIS, an Energy Conservation and Interconnection Scheme between WSN and internet based on the 6LoWPAN. 2013 IEEE 16th International Conference on Network-Based Information Systems (NBiS); 2013.
42. Oliveira LB, Aranha DF, Gouvea CP, Scott M, Camara DF, Lopez J, et al. TinyPBC: pairings for authenticated identity-based non-interactive key distribution in sensor networks. Comput Comm. 2011; 34(3):485–93.
43. Bashir MB, Abd Latiff MS, Abdulhamid SM, Loon CT. Grid-based search technique for massive academic publications. The 2014 Third ICT International Student Project Conference (ICT-ISPC2014). Thailand: IEEE Thailand; 2014. p. 175–8.
44. de Hon B, Arnold J. Discrete Green's function diakoptics - a toy problem. 2010 IEEE International Conference on Electromagnetics in Advanced Applications (ICEAA); 2010.
45. Yokokohji Y, Iida Y, Yoshikawa T. 'Toy problem' as the benchmark test for teleoperation systems. Adv Robot. 2003; 17(3):253–73.
46. Som S, Kotal A, Chatterjee A, Dey S, Palit S. A colour image encryption based on DNA coding and chaotic sequences. 2013 IEEE 1st International Conference on Emerging Trends and Applications in Computer Science (ICETACS); 2013.
47. Dowell J, Long J. Target paper: conception of the cognitive engineering design problem. Ergonomics. 1998; 41(2):126–39.
48. Garg H. Solving structural engineering design optimization problems using an artificial bee colony algorithm. J Ind Manag Optim. 2014; 10(3):777–94.
49. Alvim AC, Taillard ED. POPMUSIC for the world location-routing problem. EURO Journal on Transportation and Logistics. 2013; 2(3):231–54.
50. Sakalli A, Yesil E, Musaoglu E, Ozturk C, Dodurka MF. Heuristic bubble algorithm for a linehaul routing problem: an extension of a vehicle routing problem with pickup and delivery. 2013 IEEE 14th International Symposium on Computational Intelligence and Informatics (CINTI); 2013.
51. Paul M, Sanyal G. Task-scheduling in cloud computing using credit based assignment problem. Int J Electron Comput Sci Eng. 2011; 3(10):3426–30.
52. Frincu ME. Scheduling highly available applications on cloud environments. Future Generat Comput Syst. 2014; 32:138–53.
53. Abdulhamid SM, Abd Latiff MS, Bashir MB. On-demand grid provisioning using cloud infrastructures and related virtualization tools: a survey and taxonomy. Int J Adv Stud Comput Sci Eng. 2014; 3(1):49–59.
54. Klincewicz JG. Using GRASP to solve the generalised graph colouring problem with application to cell site assignment. Int J Mobile Netw Des Innovat. 2012; 4(3):148–56.
55. Lewis R, Thompson J. On the application of graph colouring techniques in round-robin sports scheduling. Computers and Operations Research. 2011; 38(1):190–204.
56. Hartmann S, Briskorn D. A survey of variants and extensions of the resource-constrained project scheduling problem. Eur J Oper Res. 2010; 207(1):1–14.
57. Fang C, Wang L. An effective shuffled frog-leaping algorithm for resource-constrained project scheduling problem. Computers and Operations Research. 2012; 39(5):890–901.
58. Zamani R. Integrating iterative crossover capability in orthogonal neighbourhoods for scheduling resource-constrained projects. Evol Comput. 2013; 21(2):341–60.
59. Deblaere F, Demeulemeester E, Herroelen W. Generating proactive execution policies for resource-constrained

projects with uncertain activity durations. FBE Research Report KBI_1006. 2010:1–29.
60. Cano A, Zafra A, Ventura S. Speeding up multiple instance learning classification rules on GPUs. Knowl Inform Syst. 2014:1–19.
61. Abdulhamid SM, Waziri VO, Idris L. The Application of Decidability Theory to Identify Similar Computer Networks. IUP Journal of Computational Mathematics. 2011; 4(2):26–36.
62. De Campos CP, Ji Q. Efficient structure learning of Bayesian networks using constraints. The Journal of Machine Learning Research. 2011; 12:663–89.
63. Subramanian K, Savitha R, Suresh S. A complex-valued neuro-fuzzy inference system and its learning mechanism. Neurocomputing. 2014; 123:110–20.
64. Saifullah HM, Stutzle T. Tabu search vs. simulated annealing as a function of the size of quadratic assignment problem instances. Computers and Operations Research. 2014; 43:286–91.
65. Yang X, Li X, Gao Z, Wang H, Tang T. A cooperative scheduling model for timetable optimization in subway systems. IEEE Transactions on Intelligent Transportation Systems. 2013; 14(1):438–47.
66. Zhong J-H, Shen M, Zhang J, Chung H-H, Shi Y-H, Li Y. A differential evolution algorithm with dual populations for solving periodic railway timetable scheduling problem. IEEE Transactions on Evolutionary Computation. 2013; 17(4):512–27.
67. Gharehchopogh FS, Arjang H. Survey and Taxonomy of Leader Election Algorithms in Distributed Systems. Indian Journal of Science and Technology. 2014; 7(6):815–30.
68. Qing L, Odaka T, Kuroiwa J, Shirai H, Ogura H. A New Artificial Fish Swarm Algorithm for the Multiple Knapsack Problem. IEICE TRANSACTIONS on Information and Systems. 2014; 97(3):455–68.
69. Sarac T, Sipahioglu A. Generalized quadratic multiple knapsack problem and two solution approaches. Computers and Operations Research. 2014; 43:78–89.
70. Garcia-Martinez C, Rodriguez F, Lozano M. Tabu-enhanced iterated greedy algorithm: case study in the quadratic multiple knapsack problem. Eur J Oper Res. 2014; 232(3):454–63.
71. Andrade DV, Resende MG, Werneck RF. Fast local search for the maximum independent set problem. Journal of Heuristics. 2012; 18(4):525–47.
72. Ribeiro GM, Mauri GR, Lorena LAN. A lagrangean decomposition for the maximum independent set problem applied to map labelling. Operational Research. 2011; 11(3):229–43.
73. Wang Z, Tan J, Zhu L, Huang W. Solving the Maximum Independent Set Problem based on Molecule Parallel Supercomputing. Appl Math. 2014; 8(5):2361–6.
74. Maslov E, Batsyn M, Pardalos PM. Speeding up MCS Algorithm for the Maximum Clique Problem with ILS Heuristic and Other Enhancements. Models, Algorithms, and Technologies for Network Analysis. Springer; 2013. p. 93–9.
75. Pattabiraman B, Patwary MMA, Gebremedhin AH, Liao W-k, Choudhary A. Fast algorithms for the maximum clique problem on massive sparse graphs. Algorithms and Models for the Web Graph. Springer; 2013. p. 156–69.
76. Garg H, Sharma S. Multi-objective reliability-redundancy allocation problem using particle swarm optimization. Comput Ind Eng. 2013; 64(1):247–55.
77. Yeh W-C. Orthogonal simplified swarm optimization for the series–parallel redundancy allocation problem with a mix of components. Knowl Base Syst. 2014; 64:1–12.
78. Gohareh MM, Karimi B, Khademian M. A simulation-optimization approach for open-shop scheduling problem with random process times. Int J Adv Manuf Tech. 2014; 70(5–8):821–31.
79. Ahmadizar F, Rabanimotlagh A. Group shop scheduling with uncertain data and a general cost objective. Int J Adv Manuf Tech. 2014; 70(5–8):1313–22.
80. Bezerra LC, Lopez-Ibanez M, Stutzle T. Deconstructing multi-objective evolutionary algorithms: an iterative analysis on the permutation flow-shop problem. 2014.
81. Cacchiani V, Hemmelmayr V, Tricoire F. A set-covering based heuristic algorithm for the periodic vehicle routing problem. Discrete Appl Math. 2014; 163:53–64.
82. Maher B, Albrecht AA, Loomes M, Yang X-S, Steinhofel K. A Firefly-inspired method for protein structure prediction in lattice models. Biomolecules. 2014; 4(1):56–75.
83. Hajmohammad M, Salari M, Hashemi S, Esfe MH. Optimization of Stacking Sequence of Composite Laminates for Optimizing Buckling Load by Neural Network and Genetic Algorithm. Indian Journal of Science and Technology. 2013; 6(8):5070–7.
84. Blum C. Ant colony optimization: Introduction and recent trends. Physics of Life reviews. 2005; 2(4):353–73.